\documentclass[sigconf]{acmart}

\usepackage{booktabs} 
\usepackage{subfigure}
\usepackage[vlined,algoruled,commentsnumbered]{algorithm2e}
\usepackage{amsmath}
\usepackage{bm}
\usepackage{multicol}
\usepackage{multirow}
\usepackage{soul}

\newcommand{\res} {\texttt{resistance2}}
\newcommand{\magman} {\texttt{magman}}
\newcommand{\pressure} {\texttt{pressure}}

\newcommand{\base} {\texttt{baseSNGP}}
\newcommand{\msngp} {\texttt{mSNGP}}
\newcommand{\msngpls} {\texttt{mSNGP-ls}}

\newcommand{\nbsucc} {\texttt{succ}}
\newcommand{\mse} {$\texttt{MSE}_\texttt{train}$}
\newcommand{\msetest} {$\texttt{MSE}_\texttt{test}$}
\newcommand{\mseref} {$\texttt{MSE}_\texttt{ref}$}

\newcommand{\viol} {\texttt{violation}}

\newcommand{\basemodel} {\emph{reference model}}
\newcommand{\basesngp} {\emph{base SNGP}}
\def\Data{D}
\def\ConData{C}
\newcommand{\cconstr} {$C_{c}$}
\newcommand{\ctrain} {$C_{t}$}
\newcommand{\testdata} {$D_{test}$}

\let\oldnl\nl
\newcommand{\nonl}{\renewcommand{\nl}{\let\nl\oldnl}}

\newlength\lenKwIn
\newcommand\myKwIn[1]{%
  \settowidth\lenKwIn{\KwIn{}}%
  \setlength\hangindent{\lenKwIn}%
  \nonl\hspace*{\lenKwIn}#1\\}

\newlength\lenKwOut

\SetKwData{ms}{$M$}           
\SetKwData{period}{$PERIOD$}  
\SetKwData{S}{$\mathcal{S}$}  
\SetKwData{pop}{$population$}
\SetKwData{interpop}{$interPop$}
\SetKwData{gen}{$generation$}
\SetKwData{maxgens}{$MAXGENS$}
\SetKwData{maxiters}{$MAXITERS$}
\SetKwData{lsiters}{$LS\_ITERS$}
\SetKwData{parent}{$parent$}
\SetKwData{child}{$child$}
\SetKwData{temp}{$temp$}
\SetKwFunction{abs}{abs}
\SetKwFunction{init}{init}
\SetKwFunction{appmut}{applyMutations}
\SetKwFunction{calcBeta}{calculateBetaCoeffs}
\SetKwFunction{evaltrain}{calculateC\_train}
\SetKwFunction{checkconstr}{calculateC\_constraint}
\SetKwFunction{size}{size}
\SetKwFunction{select}{selectModel}
\SetKwFunction{cp}{clone}
\SetKwFunction{dominates}{dominates}
\SetKwFunction{add}{add}
\SetKwFunction{merge}{NSGAII\_merge}
\SetKwFunction{getextreme}{getExtremeModel}
\SetKwFunction{getkmeans}{getClusterRepresentatives}
\SetKwFunction{getNondominated}{getNondominatedModels}
\SetKwFunction{update}{update}

\SetKwData{popa}{$popA$}  
\SetKwData{popb}{$popB$}
\SetKwData{fpop}{$finalPop$}
\SetKwData{tpop}{$tempPop$}
\SetKwData{flag}{$flag$}
\SetKwData{F}{$\mathcal{F}$}  
\SetKwFunction{findfronts}{findNondominatedFronts}
\SetKwFunction{getfront}{getFront}
\SetKwFunction{sort}{sortByCrowding}
\SetKwFunction{removeb}{removeBest}
\SetKwFunction{removew}{removeWorst}

\setcopyright{rightsretained}

\copyrightyear{2020}
\acmYear{2020}
\setcopyright{acmcopyright}\acmConference[GECCO '20]{Genetic and Evolutionary Computation Conference}{July 8--12, 2020}{Canc\'{u}n, Mexico}
\acmBooktitle{Genetic and Evolutionary Computation Conference (GECCO '20), July 8--12, 2020, Canc\'{u}n, Mexico}
\acmPrice{15.00}
\acmDOI{10.1145/3377930.3390152}
\acmISBN{978-1-4503-7128-5/20/07}

\begin{document}
\title{Symbolic Regression Driven by Training Data and Prior Knowledge}

\author{Ji\v{r}\'{i} Kubal\'{i}k}
\affiliation{%
  \institution{Czech Institute of Informatics, Robotics and Cybernetics\\
  Czech Technical University in Prague\\
  16000 Prague, Czech Republic}
}
\email{jiri.kubalik@cvut.cz}

\author{Erik Derner}
\affiliation{%
  \institution{Czech Institute of Informatics, Robotics and Cybernetics\\
  and\\
  Department of Control Engineering\\Faculty of Electrical Engineering\\
  Czech Technical University in Prague
  16000 Prague, Czech Republic}
}
\email{erik.derner@cvut.cz}

\author{Robert Babu\v{s}ka}
\affiliation{%
  \institution{Cognitive Robotics\\
  Delft University of Technology\\
  Delft, 2628 CD, The Netherlands\\
  and}
}
\affiliation{%
  \institution{Czech Institute of Informatics, Robotics and Cybernetics\\
  Czech Technical University in Prague\\
  16000 Prague, Czech Republic}
}
\email{r.babuska@tudelft.nl}

\begin{abstract}
In symbolic regression, the search for analytic models is typically driven purely by the prediction error observed on the training data samples. However, when the data samples do not sufficiently cover the input space, the prediction error does not provide sufficient guidance toward desired models. Standard symbolic regression techniques then yield models that are partially incorrect, for instance, in terms of their steady-state characteristics or local behavior. If these properties were considered already during the search process, more accurate and relevant models could be produced. We propose a multi-objective symbolic regression approach that is driven by both the training data and the prior knowledge of the properties the desired model should manifest. The properties given in the form of formal constraints are internally represented by a set of discrete data samples on which candidate models are exactly checked. The proposed approach was experimentally evaluated on three test problems with results clearly demonstrating its capability to evolve realistic models that fit the training data well while complying with the prior knowledge of the desired model characteristics at the same time.
It outperforms standard symbolic regression by several orders of magnitude in terms of the mean squared deviation from a reference model.
\end{abstract}

%
%



\maketitle

\section{Introduction}

Many model-learning approaches have been described in the literature: time-varying linear models \cite{a,b}, Gaussian processes and other probabilistic models \cite{c,d}, deep neural networks \cite{e,f} or local linear regression \cite{h}. All these approaches suffer from drawbacks induced by the use of the specific approximation technique, such as a large number of parameters (deep neural networks), local nature of the approximator (local linear regression), computational complexity (Gaussian process), etc.
Symbolic regression (SR) is an approach that generates models in the form of analytic equations that can be constructed by using even very small training data sets. SR has been used in nonlinear data-driven modeling with quite impressive results \cite{Schmidt2009, VLADISLAVLEVA2013, Staelens2013,Alibekov16-CDC,Derner18-ICRA,Derner18-IROS,Alibekov18}.

In standard SR, the search for analytic models is driven purely by the prediction error observed on the training data samples. However, the training data may not provide a sufficient guidance towards desired models, for instance, when the data set does not sufficiently cover the input space or even when some parts of the input space are completely omitted in the data set. SR techniques then yield models that are partially incorrect, for instance, in terms of their steady-state characteristics or local behavior.
On the other hand, some information about the desired properties of the modelled system is often available. If these properties were considered already during the search process, more accurate and relevant models could be produced.

There are very few SR approaches in the literature that take into account information about the model sought other than just the minimum training error. Perhaps the most promising and the most relevant is the Counterexample-Driven Symbolic Regression \cite{Bladek_2019}, where Counterexample-Driven Genetic Programming \cite{Krawiec_2017} is used to synthesize regression models that not only comply with the training data set, but also meet formal constraints imposed on the model. A Satisfiability Modulo Theories (SMT) solver is used to verify whether a given model meets the formal specification. However, this method suffers from several deficiencies. Queries to the SMT solver are computationally very costly. Only a limited function set $\{+, -, *, /\}$ can be used since transcendental functions are not fully supported by contemporary SMT solvers. 
There are also SMT solvers that can handle functions like \emph{sine} and \emph{log}, however, they are very computationally expensive for larger models. Moreover, general nonlinear inequalities over the real numbers represent a non-decidable problem, therefore, delta-decidability is used instead. This means, one has to supply the solvers with a proper value of the delta precision parameter. In the end, the solver can return both false positive and false negative answers. 

In this paper, we propose a multi-objective SR approach that is driven by the training data as well as by the prior knowledge on the desired properties the model should exhibit.
Various types of constraints can be used such as the monotonicity of the model's output on a given interval, odd symmetry of the model, symmetry w.r.t. the input variables, steady-state characteristics of the model, etc. The properties, given in the form of formal constraints, are internally represented by a set of discrete constraint samples on which the validity of candidate models is checked.
Both aspects of model performance are treated with equal importance. Consequently, the method produces models that fit the training data as well as possible while complying with the prior knowledge of the desired model characteristics at the same time.

We use a variant of Single Node Genetic Programming (SNGP) \cite{Jackson2012,Kubalik17-IJCCI} that generates models in the form of a linear combination of possibly nonlinear features. It has been shown that SR methods producing such compound models outperform SR methods generating single tree models \cite{Arnaldo2014,Arnaldo2015,Searson2014,Kubalik2016-TCCI}.
In the standard version of the SNGP algorithm, the coefficients of the linear model are estimated using least squares. In the proposed multi-objective SNGP, this may not be the most efficient way since the estimation of the coefficients is biased in the direction of models fitting well the training data. So, we propose an alternative way to derive the coefficients using a multi-objective local search procedure.

The paper is organized as follows. Section~\ref{problem_definition} defines the problem. Section ~\ref{method} describes the proposed method. In Section~\ref{experiments}, the method is experimentally evaluated on three test problems. Section~\ref{conclusions} concludes the paper and suggests topics for future research.
\section{Problem Definition}
\label{problem_definition}

We solve the problem of constructing an optimal analytic model $f \texttt{:} \mathcal{X} \rightarrow \mathrm{R}$ operating in the input space $\mathcal{X}\subset\mathrm{R}^n$ given two optimization goals:
\begin{itemize}
  \item The model fits the training data as accurately as possible.
  \item The model is consistent with the constraints imposed on the model that capture the desired model's properties.
\end{itemize}

Two data sets are used to search for the model, \emph{standard training data set} and \emph{constraint data set}.

\emph{Standard training data set} $\Data = \{d_1, \ldots, d_{m}\}$. Each training sample $d_i$ is a tuple
      \begin{equation}
        d_i = \langle \mathbf{x}_{i}, y_{i}\rangle\texttt{,}
        \label{di}
      \end{equation}
      where $\mathbf{x}_{i}\in \mathcal{X}$ is a particular input vector and $y_{i} \in R$ is the corresponding desired target value.

\emph{Constraints and constraint data set}. We assume that all constraints can be written as nonlinear inequality and equality constraints. Inequality constraints are:
      \begin{eqnarray}
            g_i^f(x) \leq 0,  \quad i = 1, \ldots, p
            \label{eq:ineq}
      \end{eqnarray}
      where function $g_i$ has a specific form for each particular type of inequality constraint and in general may have more arguments: $g_i(x_1, x_2, \ldots)$. For instance, to specify a monotonically increasing function, we can define $g_i^f(x_1,x_2) = f(x_1) - f(x_2)$ and then, when checking whether the constraint is satisfied, evaluate it for any pair of data points $x_1 \leq x_2$, $x_1,x_2 \in \mathcal{X}$. In the sequel, to avoid notational clutter we will write $g_i$ with a single argument as in (\ref{eq:ineq}).
      Inequality constraint violation for model $f$ is calculated as follows:
      \begin{equation}
         E_g^f = \sum_{i=1}^p \sum_{\forall x_\ell \in X_i^g} (\max(g_i^f(x_\ell),0))^2
         \label{e_g}
      \end{equation}
      where $X_i^g$ is a set of data points on which the violation of constraint $g_i$ is calculated.
      Analogously, equality constraints have the form:
      \begin{eqnarray}
            h_j^f(x) =0,  \quad j = 1, \ldots, q
            \label{eq:eq}
      \end{eqnarray}
      where function $h_j$ is specific to the particular type of equality constraint. Also this function
      may have more arguments. Equality constraint violation for model $f$ is calculated as follows:
      \begin{equation}
         E_h^f = \sum_{j=1}^q \sum_{\forall x_\ell \in X_j^h} (h_j^f(x_\ell))^2
         \label{e_g}
      \end{equation}
      The constraint data set is given by $C = X^g\cup X^h$, with $X^g = \bigcup_{i=1}^p X_i^g$ and $X^h = \bigcup_{j=1}^q X_j^h$.

The two aforementioned optimization goals are formally defined as follows:
\begin{itemize}
  \item \ctrain\ -- minimize the mean-squared error calculated for model $f$ on the training data set $\Data$
      \begin{equation}
         C_{t} = \frac{1}{m}\sum_{i=1}^{m}(f(\mathbf{d}_i)-y_i)^2\texttt{.}
         \label{ctrain}
      \end{equation}
  \item \cconstr\ -- minimize the mean-squared error calculated for model $f$ on the constraint data set $\ConData$
      \begin{equation}
         C_{c} = \frac{E_g^f + E_h^f}{|C|}\texttt{.}
         \label{cconstr}
      \end{equation}
\end{itemize}

Importantly, both aspects of the model's performance -- i.e., its accuracy as well as its formal validity -- are treated as equally important through the optimization process.

\section{Method}
\label{method}

In this section, the proposed multi-model SR method based on the SNGP algorithm is described. Firstly, the base SNGP algorithm and its population structure for storing and operating with a single analytic model are briefly described. Then, we introduce an extended population structure that allows for operating with multiple independent models. Finally, the algorithm itself is described with the focus on the multi-objective aspect of the search process.

\subsection{Base SNGP}
\label{base_sngp}
The idea of the proposed multi-objective symbolic regression method is applicable to any population-based approach. Here, we adopt a variant of SNGP \cite{Jackson2012} and particularly the variant proposed in \cite{Kubalik17-IJCCI}. In the following text we use the term ``\emph{base SNGP}'' to refer to this algorithm.

Standard SNGP is a tree-based genetic programming (GP) technique that evolves a population of individuals, i.e. program nodes, organized in an ordered linear array structure. 
The nodes are interconnected in the left-to-right manner, meaning that a node can act as an input operand only of those nodes which are positioned to its right in the population. Thus, the population of nodes represents a whole set of tree-based programs rooted in its individual nodes. 
In the context of SR, the population starts with constant nodes and variables followed by general function nodes chosen from a set $\mathcal{F}$ of elementary functions defined by the user for the problem at hand, see Figure~\ref{fig:populations}a. 
The expression trees rooted in function nodes provide a capacity to represent complex and possibly non-linear analytic functions. 
The population is evolved through a first-improvement iterative local search procedure using a mutation operator that varies the input links of the function nodes.

An important property of the \basesngp\ is that it evolves linear-in-parameters nonlinear analytic models of the form
\begin{equation}
    f(x) = \beta_0 + \sum_{i=1}^{n_f} \beta_i \varphi_i(x)
    \label{lin_model}
\end{equation}
where the nonlinear functions $\varphi_i(x)$ are \emph{features} constructed by means of GP operations using a predefined set of elementary functions $\mathcal{F}$. 
The coefficients $\beta_i$ are not evolved using genetic operators. Instead, they are estimated using some multiple regression technique, e.g. the least squares one. 
Importantly, the whole population represents a single analytic model whose features, $\varphi_i(x)$, are rooted in so-called \emph{identity nodes}, where each identity node just refers to some non-constant-output node in the population, see Figure~\ref{fig:populations}a.

The complexity of evolved analytic models is constrained by two user-defined parameters: $n_f$ is the maximum number of features the analytic model can be composed of, and $\delta$ is the maximal depth of the feature's tree representation.

We chose for this variant of GP since it has recently been shown in \cite{Arnaldo2014,Arnaldo2015,Searson2014,Kubalik2016-TCCI} that GP methods evolving this kind of compound regression models outperform conventional GP evolving a single-tree structure representing the whole model. In particular, the \basesngp\ has been successfully used for several SR tasks from the reinforcement learning and robotics domains \citep{Alibekov16-CDC,Derner18-ICRA,Derner18-IROS,Alibekov18}.
A detailed description of the \basesngp\ is beyond the scope of this paper. For more details please refer to \citep{Kubalik17-IJCCI}.

\subsection{Multi-objective SNGP}
\label{mo_sngp}

\begin{figure}[ht]
\centerline{
    \subfigure[]{\includegraphics[width=8.0cm]{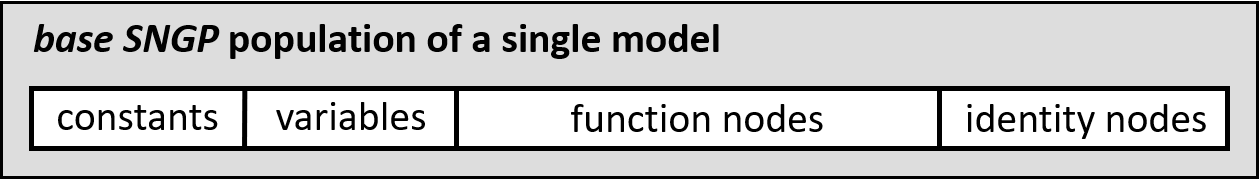}}
}
\centerline{
    \subfigure[]{\includegraphics[width=8.0cm]{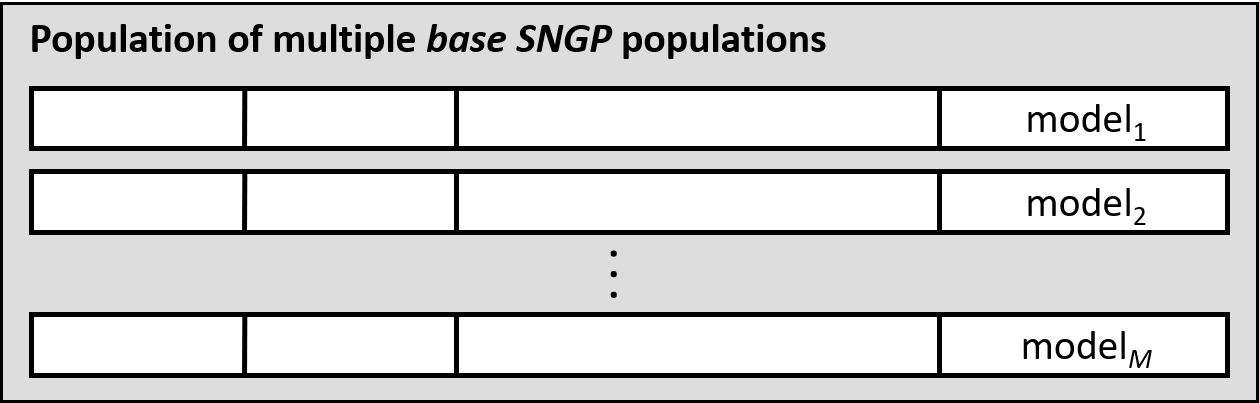}}
}
\caption{(a) Structure of the \basesngp\ population with a set of identity nodes defining features of a single model. (b) Population of models, each represented by a unique \basesngp\ population.}
\label{fig:populations}
\end{figure}

We propose a multi-objective variant of SNGP for the bi-objective SR that simultaneously optimizes both optimization criteria, \ctrain\ and \cconstr. First, we adapt the population architecture to allow for operating with a set of $M$ independent models. For this purpose, we use a set of \basesngp\ populations, each representing a unique model, see Figure~\ref{fig:populations}b. From now on, we will use the term \emph{population} in the sense of the population of models.

The proposed algorithm is based on the NSGA-II algorithm \cite{nsga02} that uses the following \emph{domination} principle: A solution $\mathbf{x}^{(1)}$ is said to dominate another solution $\mathbf{x}^{(2)}$, if $\mathbf{x}^{(1)}$ is not worse than $\mathbf{x}^{(2)}$ in any objective and $\mathbf{x}^{(1)}$ is strictly better than $\mathbf{x}^{(2)}$ in at least one objective.

The outline of the multi-objective SNGP algorithm is shown in Algorithm~\ref{alg:moea}. It starts with a random initialization of the population of models, \pop. Each model is first assigned its coefficients $\mathbf{\beta}$. In the original \basesngp, these are estimated using least squares. However, this might not be the best choice when solving multi-objective SR as will be discussed later in this section. Complete models are evaluated on both data sets $\Data$ and $\ConData$. 

The algorithm then iterates through a specified number of generations, lines~\ref{alg6}--\ref{alg24}. In each generation, an intermediate population of models, \interpop, is created from models of the current \pop, lines~\ref{alg9}--\ref{alg21}. First, a parent model is selected from the \pop and its copy is assigned as the initial value to the offspring model, \child. 
A standard tournament selection uses the \emph{crowded-comparison operator} \cite{nsga02} to choose parental models to be mutated. The crowded-comparison operator takes two models and returns the one that is from the better non-dominated front or if both are from the same non-dominated front the more unique one is returned. Thus, well-performing and unique models are preferred.

The \child then undergoes a predefined number of optimization iterations, lines~\ref{alg13}--\ref{alg20}. In each iteration, the \child is mutated and its coefficients $\mathbf{\beta}$ are recalculated. The mutated model \temp is then evaluated and it becomes the \child for the next iteration if it is not dominated by the current version of \child nor by the \parent model, lines~\ref{alg19}--\ref{alg20}. Final version of the \child is added to \interpop. 

Once \interpop has been completed, it is merged with the current \pop resulting in a new version of the \pop. This is done using the NSGA-II replacement strategy that again prefers non-dominated solutions to the dominated ones and among solutions of the same non-dominated front the more unique ones are preferred. For more details refer to \cite{nsga02}. In the end of the generation, the $\ConData$ data set can optionally be updated, see Section~\ref{constr_management}. 

Finally, a set of final models is selected as the output of the run. 
Since this is a multi-objective optimization approach, the population contains a whole set of non-dominated solutions in the end. So, the question is how to choose the best solutions to be returned as the output of the run?
Definitely, the \emph{extreme} model with the best value of \cconstr\ should be in the final set of solutions. However, this may not necessarily be the most interesting one as it can do poorly in the other objective. We will demonstrate this in Section~\ref{experiments}.
On the other hand, it is very likely that the model with the best value of \ctrain\ does not belong to the most useful ones unless it coincides with the \emph{extreme} model. The rationale for it is that such a model probably over-fits the training data while ignoring the constraints imposed on the model's properties, which is not what we want to get.
So, we need to take into consideration also the high-quality trade-off solutions. Here, we take the whole set of non-dominated solutions of the final population.

\LinesNumbered
\SetKwFor{Times}{}{times do}{end}
\DontPrintSemicolon
\begin{algorithm}[t]
\fontsize{8.3}{9.6}\selectfont
\KwIn{\ms $\dots$ size of the population of models}
\myKwIn{$\ConData\ \dots$ set of constraint samples}
\myKwIn{$\Data\ \dots$ training data set}
\myKwIn{\maxgens $\dots$ maximum number of generations}
\myKwIn{\maxiters $\dots$ maximum number of iterations carried out to produce an offspring model from the parent one}
\myKwIn{\period $\dots$ number of generations between updates of $\ConData$}
\KwOut{\S $\dots$ set of final models}
\vspace{-0.5em}
\nonl\hrulefill\\
   \init{\pop}\;\nllabel{alg1}
   \For{$\forall model \in$ \pop} {\nllabel{alg2}
      $model.$\calcBeta{}    \;\nllabel{alg2b}
      $model.$\evaltrain{\Data}    \;\nllabel{alg3}
      $model.$\checkconstr{\ConData}  \;\nllabel{alg4}
   }
   \gen $\leftarrow 0$ \;\nllabel{alg5}
   \While{\gen $<$ \maxgens} {\nllabel{alg6}
      \gen $\leftarrow$ \gen + 1 \;\nllabel{alg7}
      \interpop $\leftarrow$ \texttt{\{\}}\;\nllabel{alg8}
      \While{\interpop.\size{} $<$ \ms} {\nllabel{alg9}
         \parent $\leftarrow$ \select{\pop} \;\nllabel{alg10}
         \child $\leftarrow$ \parent.\cp{} \;\nllabel{alg11}
         $i \leftarrow 0$ \;\nllabel{alg12}
         \While{$i <$ \maxiters} {\nllabel{alg13}
            $i \leftarrow i + 1$ \;\nllabel{alg14}
            \temp $\leftarrow$ \child.\cp{} \;\nllabel{alg15}
            \temp.\appmut{} \;\nllabel{alg16}
            \temp.\calcBeta{}    \;\nllabel{alg16b}
            \temp.\evaltrain{\Data}    \;\nllabel{alg17}
            \temp.\checkconstr{\ConData}  \;\nllabel{alg18}
            \If{\texttt{!}\parent.\dominates{\temp} $\wedge$ \texttt{!}\child.\dominates{\temp}} {\nllabel{alg19}
               \child $\leftarrow$ \temp \;\nllabel{alg20}
            }
         }
         \interpop.\add{\child} \;\nllabel{alg21}
      }
      \pop $\leftarrow$ \merge{\pop, \interpop} \;\nllabel{alg22}
      \If{\gen $\%$ \period $== 0$}  {\nllabel{alg23}
         $\ConData$.\update{} \;\nllabel{alg24}
      }
   }
   \S $\leftarrow$ \pop.\getNondominated{} \;\nllabel{alg25}
\KwRet{\S} \;\nllabel{alg28}
\caption{Multi-objective SNGP algorithm}
\label{alg:moea}
\end{algorithm}

\subsection{Alternative way to estimate coefficients $\mathbf{\beta}$}

As mentioned above, the coefficients $\mathbf{\beta}$ weighting the model's features in (\ref{lin_model}) are fitted using the least squares method. In particular, the coefficients are found such that the final model minimizes the sum of squared residuals over the training data set $\Data$. Thus, just the \ctrain\ objective is considered at that moment, the \cconstr\ is ignored. This means, there is no pressure towards coefficients that would make the model better in terms of \cconstr, even if it was attainable with the given set of features. Clearly, this may adversely affect the performance of the whole method. In order to remedy this issue, we propose an alternative way to calculate the coefficients so that both objectives are optimized simultaneously.

We adopt a simple local search method to tune the coefficients $\mathbf{\beta}$. It initializes the coefficients with values uniformly sampled from interval $(-1,1)$ and the performance measures of the initial model, \ctrain\ and \cconstr, are calculated. Then, it iterates for the specified number of iterations. In each iteration, the vector $\mathbf{\beta}$ is perturbed by adding values sampled from the normal distribution according to
$$\mathbf{\beta'} \leftarrow \mathbf{\beta} + \mathcal{N}(0,0.1).$$

The new vector of coefficients $\mathbf{\beta'}$ is accepted if the model using it dominates the model with the current values $\mathbf{\beta}$. Otherwise, the current vector $\mathbf{\beta}$ remains for the next iteration.

\subsection{Constraint Data Set Management}
\label{constr_management}

The content of the constraint sample set $\ConData $ is crucial for the success of the method.
Initial constraint samples, i.e., the points $\mathbf{x}$ of the input space where the constraint will be checked, are drawn randomly with a uniform distribution from the whole input space $\mathcal{X}$.
Similarly, new constraint samples that are added to $\ConData $ during the optimization process are generated at random. Note that even such a simple method can be beneficial for the optimization process as the newly added constraint samples can alter the ranking of models in the population. Such an intervention can boost the exploration towards different regions of the search space and can help to prevent the population from stagnating.

Obviously, more sophisticated sampling strategies could improve the performance of this method. However, this is out of the scope of this paper. We leave this for the future research.
\section{Experiments}
\label{experiments}

Three methods were compared:
\begin{itemize}
  \item \base\ -- \basesngp\ minimizing only the mean squared error on the training data set as described in Section~\ref{base_sngp},
  \item \msngp\ -- the proposed multi-objective SNGP using the least squares method to estimate coefficients $\beta$,
  \item \msngpls\ -- the proposed multi-objective SNGP using the local search procedure to estimate coefficients $\beta$.
\end{itemize}
The methods were experimentally evaluated on three problems:
\begin{itemize}
  \item \textbf{\res} -- This is a test problem originally proposed in \cite{Bladek_2019}. It uses a sparse set of noisy samples derived using the equivalent resistance of two resistors in parallel, $r=r_1 r_2/(r_1+r_2)$, denoted as a \basemodel. The goal is to find such a model $f(r_1,r_2)$ that fits the training data and has the same properties as the \basemodel. 
  For the sake of a unified notation, we define $x=(r_1,r_2)$ and $y = r = f(x_1,x_2)$. 
  \item \textbf{\magman} -- The magnetic manipulation system consists of an iron ball moving along a rail and an electromagnet at a static position under the rail. The goal is to find a model of the nonlinear magnetic force affecting the ball, $f(x)$, as a function of the horizontal distance, $x$, between the iron ball and the activated coil given a constant current through the coil, $i$. We use data measured on a real system and an empirical model $\tilde{f}(x)=-i c_1 x / (x^2+c_2)^3$ proposed in the literature \cite{Hurak2012} as the \basemodel. Parameters $c_1$ and $c_2$ were found empirically for the given system and this model was used to design well-performing nonlinear controllers in \cite{Damsteeg2017ModelBasedReal, Alibekov18}. 
  For this example, we define $y = f(x)$. 
  \item \textbf{\pressure} -- In this problem, highly nonlinear pressure dynamics in a laboratory fermenter is modelled. The process under consideration is a 40\,l laboratory fermenter which contains 25\,l of water. At the bottom of the fermenter, air is fed into the water at a specified flow rate which is kept at a desired value by a local mass-flow controller. The air pressure $p$ in the head space can be controlled by opening or closing an outlet valve $u$ at the top of the fermenter. The goal is to find a dynamic model $p_{k+1} = f(p_k,u_k)$. The exact form of the nonlinear target function is unknown. For more details, please refer to \cite{Babuska1998FuzzyModelingControl}. 
  For this example, we define $x=(p_k,u_k)$ and $y = f(x_1,x_2)$. 
\end{itemize}

We chose these three problems since we possess detailed knowledge of the data and they have a potential to demonstrate advantages of the proposed approach. For all the problems, the training data set $\Data$ is either very sparse or its samples are unevenly distributed in the input space. Standard symbolic regression has a very small chance of converging to an acceptable model as it is likely to over-fit the data. In addition, it is easy to visualize the models. 

Fifty independent runs were carried out with each method on each problem. 
In order to assess statistical significance of the differences among the algorithms we used the Wilcoxon rank sum test, which rejects the null hypothesis that the two compared sets are sampled from continuous distributions with equal medians at the 1\% significance level. 
In the tables, two cases are highlighted -- whether \base\ is significantly better or worse than both new methods and whether \msngp\ performance is significantly different from that of \msngpls.

The algorithms were tested with the following parameter setting:
\begin{itemize}
  \item Population size of each \basesngp\ population: 400
  \item Maximum number of features:
  \begin{itemize}
    \item \res: $n_f=3$
    \item \magman, \pressure: $n_f=5$
  \end{itemize}
  \item Maximum feature's depth:
  \begin{itemize}
    \item \res: $\delta=5$
    \item \magman, \pressure: $\delta=7$
  \end{itemize}
  \item Elementary functions:
  \begin{itemize}
    \item \res: $\mathcal{F}=\{+,\,-,\,*,\,/\}$
    \item \magman, \pressure: $\mathcal{F}=\{+,\,-,\,*,\,square,\,cube,\,sine,\,tanh\}$
  \end{itemize}
  \item Population size: $M=50$
  \item Tournament size: 3
  \item Maximum number of generations: $MAXGENS=40$
  \item Maximum number of iterations: $MAXITERS=50$
  \item Number of local search iterations: $LS\_ITERS=50$
  \item Number of generations between $\ConData$ updates: $PERIOD=2$
  \item Total number of fitness evaluations in \base:\\ $M\times MAXGENS\times MAXITERS = 10^5$.
\end{itemize}

For the \res\ problem, just elementary arithmetic operators are used and rather low-complexity models are allowed. However, such a configuration is not sufficient for more difficult problems like \magman\ and \pressure. Thus, a richer set of elementary functions as well as parameters allowing for more complex models were used for these two problems.

\subsection{Resistance2}
\label{res2}

\subsubsection{Training data}
\label{res2_train_data}
We use the data set with 10 training samples that was used in \cite{Bladek_2019}. The values of $x_1$ and $x_2$ are sampled uniformly from the interval [0.0001, 20], see Figure~\ref{fig:models_res2}. The variables as well as the target value of each training sample are disturbed with a noise randomly generated with a normal distribution $\mathcal{N}(0,0.1\sigma_X)$, where $X$ is a given variable. 
Such a noisy training data set is generated anew for each independent symbolic regression run.

\subsubsection{Constraints}
We used the following three constraints as defined in \cite{Bladek_2019}:
\begin{itemize}
  \item symmetry with respect to arguments: $f(x_1,x_2) = f(x_2,x_1)$,
  \item domain-specific constraint: $x_1=x_2 \Longrightarrow f(x_1,x_2)=\frac{x_1}{2}$,
  \item domain-specific constraint: $f(x_1,x_2)\leq x_1$, $f(x_1,x_2)\leq x_2$.
\end{itemize}

The initial set $\ConData$ contained 60 constraint samples.

\subsubsection{Performance evaluation}
We use a two-phase procedure to evaluate the obtained models. First, a model is checked whether it is ``reasonably'' close to the \basemodel\ $r(\cdot)$. For this purpose, a high-resolution grid of $200\times200$ validation points sampled in the input space $[0.0001, 20]^2$ was generated. The response values of the model on all grid points are calculated. If the maximum absolute deviation, MAD, of the model's response $f(\cdot)$ from the \basemodel\ $r(\cdot)$ is less than $\epsilon=0.1\sigma_y$ over all validation points then the model is considered as \emph{acceptable}. From each run of the multi-objective SNGP algorithm, only the acceptable model with the least MAD value is selected to the set of acceptable models.

Then, the following four performance measures are defined:
\begin{itemize}
  \item \nbsucc\ -- the number of runs, which yielded an acceptable model.
  \item \mse\ -- median \ctrain\ value over the set of acceptable models.
  \item \mseref\ -- median of the mean squared error between the model's output and the \basemodel\ on the validation points calculated over the set of acceptable models.
  \item \viol\ -- median \cconstr\ value over the set of acceptable models.
\end{itemize}

\subsubsection{Results}
\begin{figure}[b]
    \subfigure[]{\includegraphics[width=6.5cm]{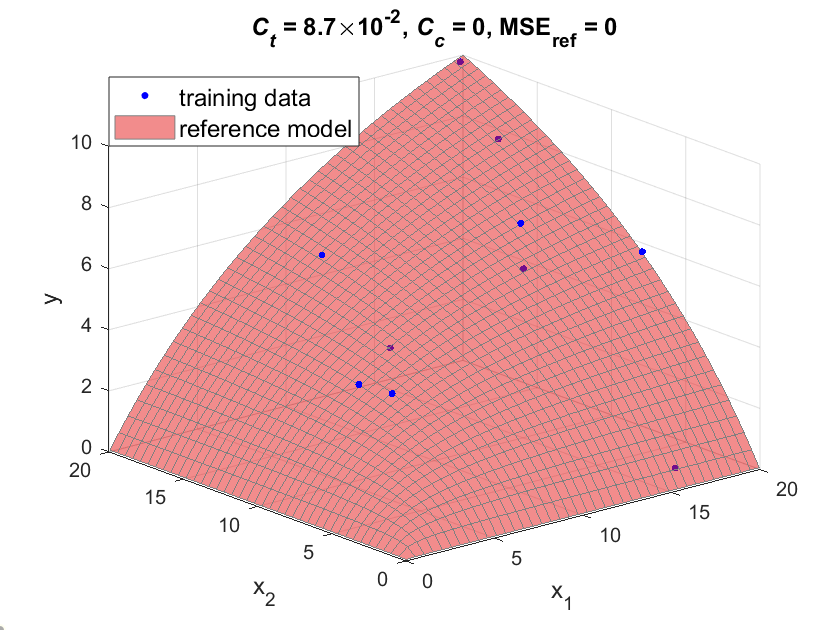}} \\
    \subfigure[]{\includegraphics[width=6.5cm]{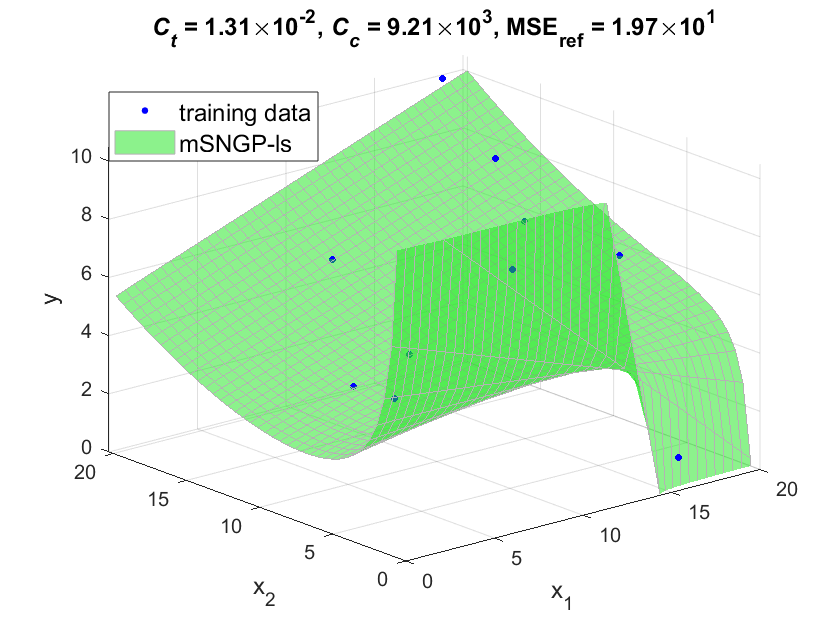}} \\
    \subfigure[]{\includegraphics[width=6.5cm]{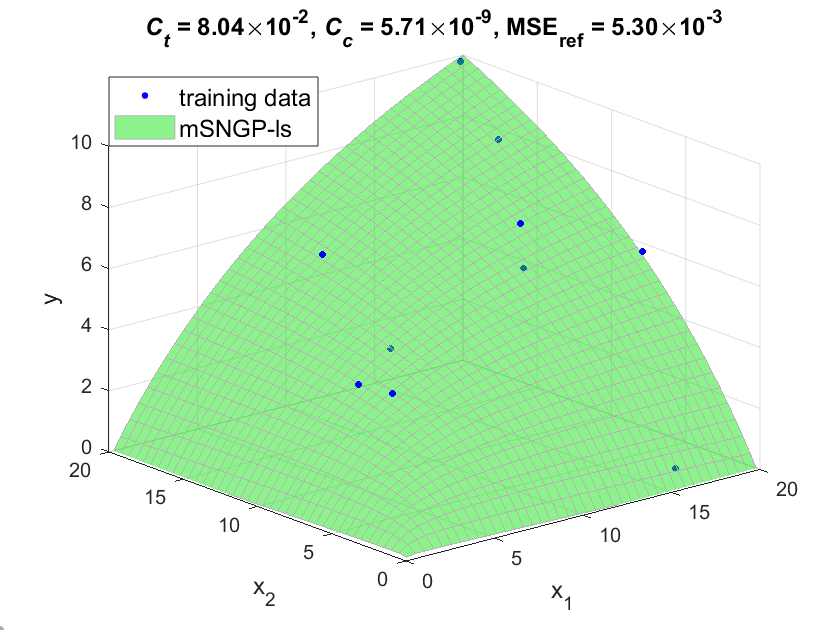}}
\caption{Models for \res\ problem. (a) The \basemodel\ $r(r_1, r_2)=r_1 r_2/(r_1+r_2)$, from which the training data were sampled. (b) An example of a model evolved with \base\ that perfectly fits $\Data$, but violates the physical law. (c) An example of a high-quality model evolved with \msngpls.}
\label{fig:models_res2}
\end{figure}

The results achieved on this problem are presented in Table~\ref{tab:res2}. The \basemodel\ and examples of the models obtained are in Figure~\ref{fig:models_res2}.

\begin{table}[]
\caption{Comparison on \res\ problem. Since the \base\ algorithm did not find any acceptable model, its \mse\, \mseref\ and \viol\ values are calculated over all fifty resulting models. \mse\ values shown in the brackets are the median \ctrain\ values achieved by the \basemodel\ on the training data sets used in the runs where the acceptable models were produced. Bold values indicate that \base\ is significantly better or worse than both \msngp\ and \msngpls.}
\label{tab:res2}
\centering
\begin{tabular}{|l|c|c|c|}
\hline
          & \base\ & \msngp\ & \msngpls\  \\
\hline
 \nbsucc\  & 0\,/\,50              & 8\,/\,50              & 16\,/\,50    \\
 \mse\     & $\bm{3.6 \times 10^{-3}}$  & 0.078 (0.11)          & 0.14 (0.15) \\
 \mseref\  & $\bm{1.5 \times 10^{3}}$   & $1.7 \times 10^{-3}$  & $1.6 \times 10^{-4}$ \\
 \viol\    & $\bm{2.3 \times 10^{3}}$   & $4.5 \times 10^{-4}$  & $1.1 \times 10^{-6}$ \\
\hline
\end{tabular}
\end{table}

\subsection{Magman}
\label{res2_train_data}

\subsubsection{Training data}
The portion of the input space of interest spans over the interval $-0.075\,\text{m} \leq x \leq 0.075\,\text{m}$. However, only its small part, [-0.027\,m, 0.027\,m], is covered by the 858 data samples collected for this task, see Figure~\ref{fig:models_magman}. The data were measured on a real system \cite{Damsteeg2017ModelBasedReal}. The whole data set was split into the training and test data sets, $\Data$ and \testdata, in the ratio 7:3.
Properties of the model sought outside the sampled interval are specified purely by the additional constraints.

\subsubsection{Constraints}
The following constraints were defined for the \magman\ problem. The model sought is an odd function that is positive on the interval $[-0.075,0]$ and negative on the interval $[0,0.075]$. Furthermore, it is monotonically increasing on the intervals $[-0.075,-0.008]$ and $[0.008, 0.075]$ and monotonically decreasing on the interval $[-0.008,0.008]$. Finally, we define the exact output value $f(0)=0$ and two exact output values at the boundary points of the input space as $f(-0.075)=1\times10^{-3}$ and $f(0.075)=-1\times10^{-3}$. 

The initial set $\ConData$ contained 90 constraint samples.
An example of a constraint sample of the decreasing monotonicity constraint is
$$x = \langle x_1,x_2\rangle\texttt{, } x_1, x_2\in[-0.01, 0.01]\wedge 0 < x_2-x_1<\eta,$$
and the constraint violation is calculated according to (\ref{e_g}) as $$\max(f(x_2)-f(x_1),0)^2.$$
This means, two distance values, $y_1$ and $y_2$, are sampled from a narrow interval of size $\eta$. Here, $\eta=0.0001$ was used.

\subsubsection{Performance evaluation}
\begin{figure}[]
    \subfigure[]{\includegraphics[width=6.5cm]{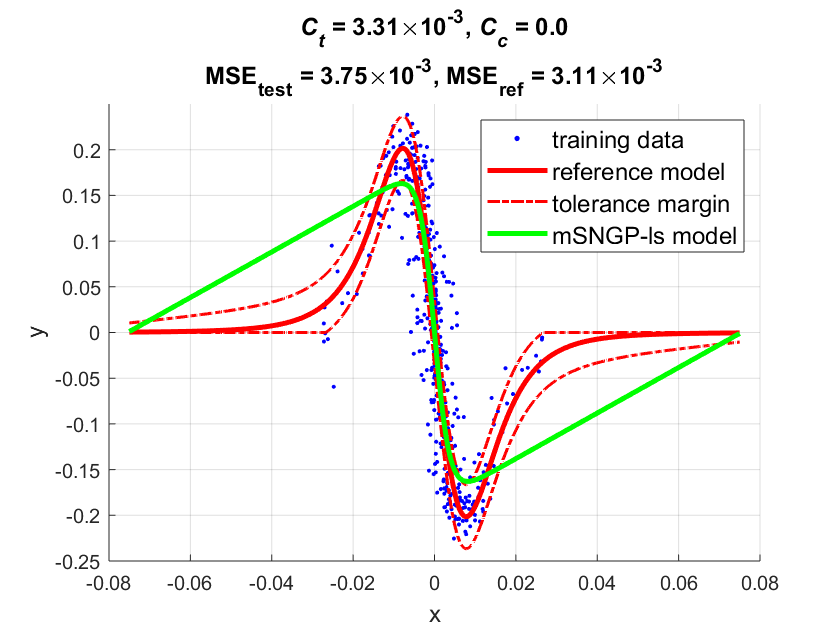}} \\
    \subfigure[]{\includegraphics[width=6.5cm]{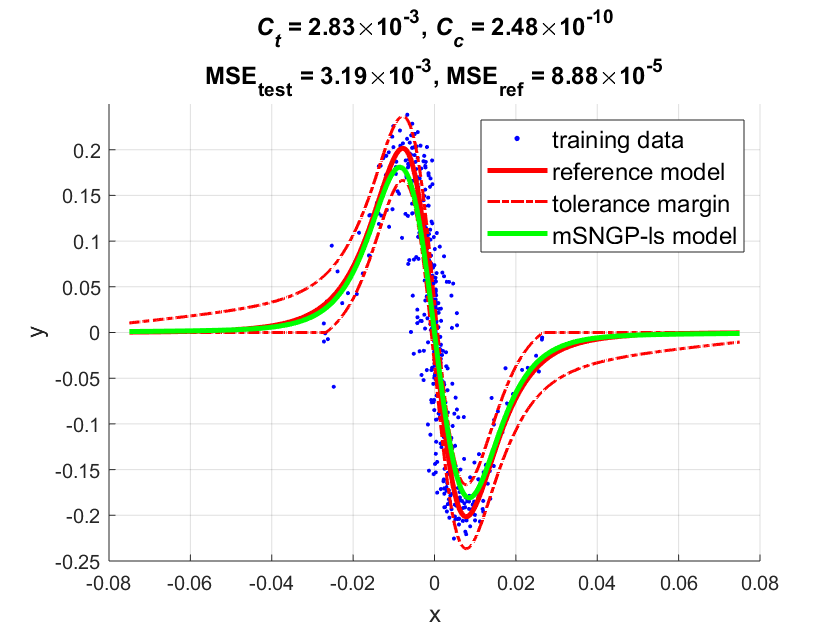}}
\caption{Models for \magman\ problem. (a) A trivial model perfectly satisfying constraints, but poorly fitting the training
data. (b) An example of well-performing model.}
\label{fig:models_magman}
\end{figure}

Similarly to the \res\ problem, the models are evaluated in two steps. First, we use 30000 validation points evenly sampled from the whole input domain [-0.075\,m, 0.075\,m] to check whether the model output lies within a tolerance margin around the \basemodel, see Figure~\ref{fig:models_magman}. 
If so, the model is considered acceptable and its mean squared deviation MSD from the \basemodel\ is calculated. From each run of the multi-objective SNGP algorithm, only the acceptable model with the least MSD value is selected to the set of acceptable models. Then, the \nbsucc, \mse, \mseref\ and \viol\ performance values are calculated in the same way as for \res\ problem. 
In addition, a \msetest\ performance measure is calculated as a median of the mean squared error on \testdata\ over the set of acceptable models. Note, the scope of the model's validation on the \testdata\ is limited since it applies only to the portion of the input space that was covered by the data.

\begin{table}[]
\caption{Comparison on \magman\ problem. The values presented in column \base\ are calculated over all fifty resulting models. Bold values in the first column indicate that \base\ is significantly better or worse than both \msngp\ and \msngpls. Bold values in the second and third column indicate that the respective method is significantly better than the other proposed method.}
\label{tab:magman}
\centering
\begin{tabular}{|l|c|c|c|}
\hline
           & \base\ & \msngp\ & \msngpls\ \\
\hline
 \nbsucc\  & 0\,/\,50                   & 15\,/\,50             & 27\,/\,50      \\
 \mse\     & $\bm{2.78 \times 10^{-3}}$ & $\bm{2.80\times 10^{-3}}$  & $2.84 \times 10^{-3}$ \\
 \msetest\ & $3.14 \times 10^{-3}$      & $\bm{3.14 \times 10^{-3}}$ & $3.22 \times 10^{-3}$ \\
 \mseref\  & \textbf{12.8}              & $1.38 \times 10^{-4}$ & $\bm{1.03 \times 10^{-4}}$ \\
 \viol\    & \textbf{5.6}               & $7.3 \times 10^{-6}$  & $\bm{2.7 \times 10^{-9}}$  \\
\hline
\end{tabular}
\end{table}

\subsubsection{Results}
The results achieved on this problem are presented in Table~\ref{tab:magman}. Figure~\ref{fig:models_magman} shows an example of a trivial model that perfectly satisfies the constraints, but poorly fits the training data and an example of an overall well-performing model.

\subsection{Pressure}

\subsubsection{Training data}
A set of 756 data samples unevenly distributed in the input space $[1,2]\times[0,100]$ were measured on the real system, see Figure~\ref{fig:models_pressure}. The data were split into the training and test data sets, $\Data$ and \testdata, in the ratio 7:3.

\subsubsection{Constraints}
Two types of constraints were defined for this problem. The model sought is monotonically increasing w.r.t. both inputs on the whole input space domain. The model's output is bounded in the interval [1, 2.2]. The initial set $\ConData$ contained 80 constraint samples.

\subsubsection{Performance evaluation}
In this case, we use a grid of $200\times200$ validation points evenly sampled from the input space $[1,2]\times[0,100]$. Models that have zero violation on all of those validation points are considered acceptable. For each run, the model with the least \ctrain\ is chosen. Only \nbsucc, \mse\ and \msetest\ values are calculated since \viol\ is by definition zero for all acceptable models and \mseref\ can not be calculated since we do not have any reference model, neither theoretical nor empirical one.

\subsubsection{Results}
\begin{figure}[]
    \subfigure[]{\includegraphics[width=6.4cm]{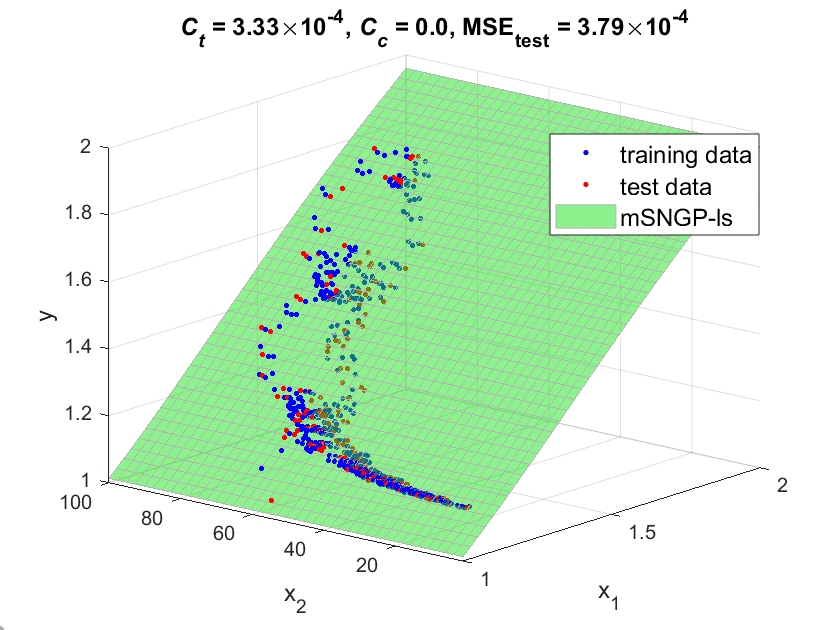}}
    \subfigure[]{\includegraphics[width=6.4cm]{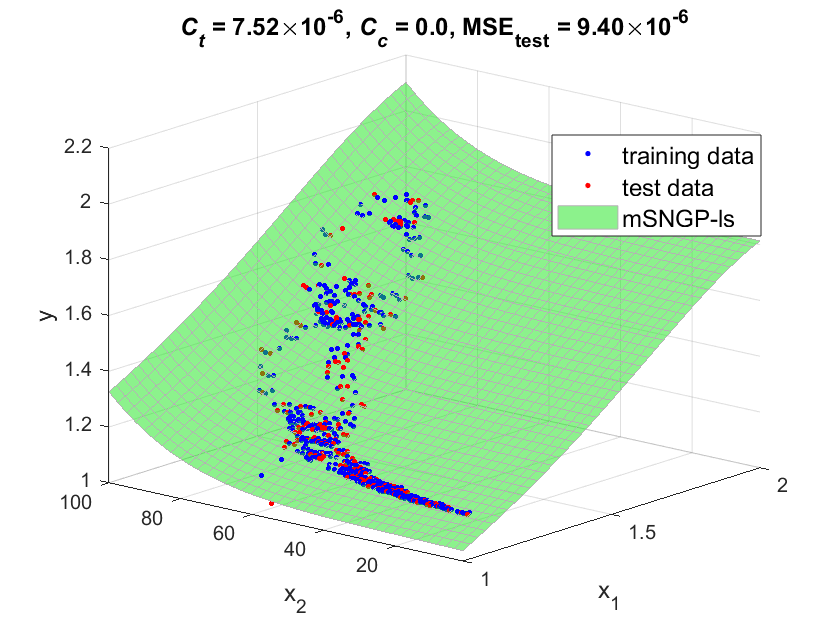}}
\caption{Models for \pressure\ problem. (a) Trivial model perfectly satisfying constraints, but poorly fitting the training data. (b) An example of an overall well-performing model.}
\label{fig:models_pressure}
\end{figure}

The results achieved on this problem are presented in Table~\ref{tab:pressure}. 
Figure~\ref{fig:models_pressure} shows an example of a trivial model that perfectly satisfies the constraints, but poorly fits the training data and an example of an overall well-performing model.

\begin{table}[]
\caption{Comparison on \pressure\ problem. The bold value in the first column indicates that \base\ is significantly better than both \msngp\ and \msngpls. Bold values in the second column indicate that \msngp\ is significantly better than \msngpls.}
\label{tab:pressure}
\centering
\begin{tabular}{|l|c|c|c|}
\hline
          & \base\ & \msngp\ & \msngpls\ \\
\hline
 \nbsucc\  & 13\,/\,50                  & 42\,/\,50                  & 49\,/\,50      \\
 \mse\     & $\bm{4.43 \times 10^{-6}}$ & $\bm{4.51 \times 10^{-6}}$ & $7.28 \times 10^{-6}$  \\
 \msetest\ & $8.83 \times 10^{-6}$      & $\bm{7.83 \times 10^{-6}}$ & $1.28 \times 10^{-5}$ \\
\hline
\end{tabular}
\end{table}

\subsection{Discussion}

\subsubsection{\res}
\base\ was not able to find any acceptable model in the fifty independent runs. The models it produces fit the training data very well, but they largely violate the constraints, as illustrated in Figure~\ref{fig:models_res2}b.
The proposed multi-objective method already finds acceptable models, an example is in Figure~\ref{fig:models_res2}c. However, none of the acceptable models found has zero \viol, none of them is an identical version of the \basemodel. Our hypothesis is that SNGP searches the space of excessively complex models and misses the simpler ones. Among these models, it is hard to find those that perfectly satisfy the equality constraints. This issue can be resolved adjusting the algorithm to search for models of varying complexity. We leave this for the future work.

In addition, both \msngp\ and \msngpls\ outperform \base\ by six orders of magnitude in terms of the \mseref.
Interestingly, both \msngp\ and \msngpls, achieve better \ctrain\ than the \basemodel\ as illustrated for one particular run in Figures~\ref{fig:models_res2}a and \ref{fig:models_res2}c. In 14 out of the 16 cases \msngpls\ found an acceptable model with the \ctrain\ value better than that of the \basemodel.

An important observation is that \msngpls\ outperforms \msngp\ in terms of all four performance indicators, though the p-values ranging from 0.022 to 0.071 returned by the rank sum test do not suggest that the observed differences are statistically significant. This can be attributed to very small size of tested samples, 8 and 16. 

\subsubsection{\magman}
The \base\ method finds the most precise models w.r.t. \ctrain, but they are effectively useless due to large constraint violation. It generated no acceptable model.
The proposed methods can find acceptable models, \msngpls\ is better than \msngp. 
Both \msngp\ and \msngpls\ outperform \base\ by five orders of magnitude in terms of the \mseref.
Similarly to \res, it is hard to find a model perfectly satisfying the constraints defined for this problem. 
Again, the small constraint violations of the acceptable models are due to the equality constraints, which is no problem, as the purpose of these constraints is to force the function asymptotically approach zero, rather than attain an exact value.
Interestingly, the acceptable models are very close to the empirical one in the regions of the input space where only the constraints were specified, see Figure~\ref{fig:models_magman}b.

As mentioned in Section~\ref{mo_sngp}, all non-dominated models are considered as the output of the \msngp\ and \msngpls\ run. The reason is that the models with the best \ctrain\ or \cconstr\ are often useless. A typical example of the model with zero constraint violation is in Figure~\ref{fig:models_magman}a. Clearly, the model is bad as it does not fit the training data well.

\subsubsection{\pressure}
This is the only problem where the \base\ method succeeded in finding acceptable models. Again, both variants of the proposed approach find an acceptable model significantly more frequently than \base, with \msngpls\ having higher success rate than \msngp.

\subsubsection{Constraint samples}
One thing that still remains an open issue is that even when a model successfully passes all constraint samples, the model may not necessarily be valid on the whole constraint domain. To increase the efficacy of the validity checks, the right constraint samples, i.e., the most informative ones, should be used during the whole run. 
In general, the ability to continuously generate arbitrary constraint samples has a great potential to direct the search to better models on the fly and it is something that should effectively be utilized.
Various strategies can be used to attain this goal. For example, new samples of a given constraint can be generated in the vicinity of the sample that has been found hard to be satisfied by models in the current population. Similarly, a set of candidate constraint samples can be randomly generated and the one with the highest fail ratio over the models in the current population is chosen. This is in accordance with the observation presented in \cite{Bladek_2019} that counterexamples are more beneficial for evolving correct models than just random samples. New samples can also be generated so that the coverage of the constraint domain increases as much as possible.

\section{Conclusions}
\label{conclusions}

We proposed a new multi-objective symbolic regression method where both aspects of the model performance, the model's accuracy as well as its formal validity, are treated equally. 
This is a general approach that is applicable whenever some information about the desired properties of the modelled system in the form of explicit samples is available. 
The results achieved through experiments on three test problems clearly demonstrate its capability to evolve realistic models that fit well the training data while complying with the prior knowledge of the desired model characteristics at the same time. We also proposed an alternative method for estimating coefficients of the linear model. This simple yet effective local search method proved to be better than the least squares method.

An advantage of the proposed method over the validation methods based on the use of SMT solvers is that it checks the model validity on discrete samples, which is fast (even for large models) and exact. Moreover, arbitrary functions can be used to build the models. 
However, the selection of constraint samples is an open issue. We will investigate various strategies to maintain the most relevant constraint samples during the whole run. Here, we intend to take an inspiration from the field of active learning.

When evaluating the validity of a candidate model, its cumulative constraint violation is calculated over all constraint samples. Note, the violations can be of a very different scale for the different constraints. Consequently, some constraints can dominate the others within the constraint violation objective. The normalization of the constraint violations is another future research line.

\section{Acknowledgements}
This work was supported by the European Regional Development Fund under the project Robotics for Industry 4.0 (reg. no. CZ.02.1.01/\\0.0/0.0/15\_003/0000470).

\bibliographystyle{acm} 
\bibliography{gecco_2020_main}

\end{document}